\documentclass[sigconf,nonacm]{acmart}

\usepackage{booktabs}      %
\usepackage{multirow}
\usepackage{subcaption}
\usepackage{enumitem}      %
\usepackage{amsmath}
\usepackage{algorithm}
\usepackage[noend]{algpseudocode}
\usepackage{xcolor}

\usepackage{multirow}
\usepackage{subcaption}
\usepackage{graphicx}
\usepackage{algorithm}
\usepackage{algorithmicx}
\usepackage{colortbl}
\numberwithin{equation}{section}

\newcommand{\name}{{\tt HAG-PS}}

\setcopyright{acmlicensed}
\copyrightyear{2025}
\acmYear{2025}
\acmDOI{10.1145/XXXXXXX.XXXXXXX}
\acmISBN{978-1-4503-XXXX-X/25/11}
\acmConference[UrbComp '25]{The 14th International Workshop on Urban Computing
}{August 2025}{Canada}

\title{Multi-Agent Reinforcement Learning for Dynamic Mobility Resource Allocation with 
Hierarchical Adaptive Grouping  
}

\author{Farshid Nooshi and Suining He}
\affiliation{%
  \institution{
  Ubiquitous \& Urban Computing Lab \\
  University of Connecticut}
  \city{Storrs}
  \state{CT}
  \country{USA}}
\email{{farshid.nooshi,suining.he}@uconn.edu}

\renewcommand{\shortauthors}{Nooshi and He}

\renewcommand\footnotetextcopyrightpermission{}

\begin{document}

\begin{abstract}
\noindent
Allocating mobility resources
(e.g., shared bikes/e-scooters, ride-sharing vehicles)
is crucial for rebalancing
the mobility demand and supply in
the urban environments. 
We propose in this work a novel multi-agent reinforcement 
learning named 
\underline{H}ierarchical 
\underline{A}daptive 
\underline{G}rouping-based 
\underline{P}arameter 
\underline{S}haring (\name{})
for dynamic mobility resource allocation. \name{}
aims to address two important research challenges regarding multi-agent reinforcement learning for mobility resource allocation:
(1) how to dynamically and adaptively share
the mobility resource allocation policy (i.e., how to distribute mobility resources) across
agents (i.e., representing
the regional coordinators of mobility resources); 
and (2) how to achieve
memory-efficient parameter sharing in an urban-scale setting.

To address the above challenges,
we have provided following novel designs within \name{}.
To enable dynamic
and adaptive parameter sharing, we have designed
a hierarchical approach
that consists of global and local information of the 
mobility resource states (e.g., distribution
of mobility resources). 
We have developed an adaptive 
agent grouping approach 
in order to split or merge the groups of agents based on 
their relative closeness of encoded trajectories (i.e., states, actions, and rewards). 
We have designed 
a learnable identity (ID) embeddings
to enable agent specialization
beyond simple parameter copy. 
We have performed extensive
experimental studies based on 
real-world NYC bike sharing data (a total of more than 1.2 million trips),  
and demonstrated 
the superior performance (e.g., improved bike availability) of
\name{} compared with 
other baseline approaches.

\end{abstract}


\settopmatter{printfolios=true}

\maketitle

\section{Introduction}\label{sec:intro}

Dynamic allocation of urban mobility resources, such as the shared bikes 
\cite{hu2021dynamic}, e-scooters 
\cite{chu2023dynamic}, and ride-sharing vehicles 
\cite{bei2018algorithms,HeS19_STAP}, 
is crucial for enhancing the  operational efficiency of urban mobility systems and satisfying the mobility needs of various communities. 
Due to the complex city environments and varying mobility needs, how to adaptively rebalance the demands and supplies is crucial for the success of 
allocation. 
Among various approaches explored to support 
city-scale rebalancing, 
multi-agent reinforcement learning (MARL) 
has been explored due to its adaptivity and scalability. 
By considering  
coordinators 
(e.g., ride sharing
drivers, bike sharing relocators) as agents,
MARL can serve as the resource distribution engine, observing 
the mobility resources and environment (states), and dynamically 
allocate the mobility resources (actions) through strategically (policy) coordinating 
these agents' behaviors.     

Despite the prior results, two major challenges remain
before the MARL 
can be deployed 
for mobility resource allocation 
and demand-supply rebalancing.
\textit{First}, how can we dynamically and adaptively share 
the mobility resource 
allocation policy (e.g., when and where
the available mobility resources
should be re-allocated) across different agents that are coordinating?
One may consider 
grouping the agents and respectively
instantiating a set of policy networks for each agent group~\cite{wang2024dyps}. 
Such a grouping method, however, may not provide a principled mechanism on the group number and their sizes. Therefore,
sub-optimal performance may be 
achieved given the dynamic urban mobility demands and supplies. 
\textit{Second},
how can we achieve 
scalable and memory-efficient parameter sharing  
in a city-wide setting?
Learning the mobility 
allocation policy for every agent 
is not feasible. 
On the other hand, a shared mobility resource allocation policy may not necessarily capture heterogeneous roles or location‑specific specialization of the agents. The performance of MARL may hence deteriorate given the complex urban environment with varying mobility demands and supplies~\cite{yang2018meanfield}.

To overcome the above-mentioned challenges, 
we propose in this work a novel multi-agent reinforcement 
learning named \name{}, 
\underline{H}ierarchical 
\underline{A}daptive 
\underline{G}rouping-based 
\underline{P}arameter 
\underline{S}haring 
for dynamic and adaptive mobility resource allocation.
Toward development of \name{},
we have made the following
contributions:
\begin{itemize}
    \item \textit{Hierarchical 
    Parameter Sharing with Scalable 
    and Memory-Efficient Designs}:
    We have designed and developed
    a hierarchical parameter sharing mechanism for MARL. Our mechanism
    consists of global and local information of the 
    mobility resource states (e.g., distribution
    of mobility resources across different regions). This way, 
    our \name{} can enable dynamic
    and adaptive parameter sharing in a city-wide setting. 
    \item 
    \textit{Adaptive Grouping of Coordinating Agents}:
    We have developed an adaptive parameter
    budget-capped
    agent grouping approach 
    in order to split or merge 
    the groups of agents based on 
    their relative closeness of 
    encoded trajectories (i.e., states, actions, and rewards). 
    We have designed 
    a learnable identity (ID) embeddings
    to enable agent specialization
    beyond simple parameter copy. 
    Through these measures, we enhance 
    the model adaptivity in dynamic mobility resource allocation. 
    \item \textit{Extensive Emulation Studies based on 
    Bike Sharing Mobility Data}:
    We have performed extensive
    experimental studies based on 
    real-world NYC bike sharing data (a total of 1,232,838 trips),  
    and demonstrated 
    the superior performance (e.g., improved bike availability and rebalanced bikes) of
    \name{} compared with 
    other baseline approaches.
\end{itemize}

The rest of the paper is organized as follows.
We first present the related work
in Section~\ref{sec:related}.
Then, we discuss the concepts,
problem formulation, and core 
designs in Section~\ref{sec:problem}. 
After that, we demonstrate the preliminary results
in Section~\ref{sec:exp}, and conclude 
with future studies in Section~\ref{sec:conclude}.

\section{Related Work}
\label{sec:related}

We briefly review the 
related work in the following
two categories. 

$\bullet$~\textbf{Parameter Sharing for MARL.}
Various parameter sharing methods
have been explored in order to improve the 
memory requirements and learning efficiency
of MARL~\cite{foerster2016ldr,gupta2017pstrpo,chu2017psddpg}. 
In order to enhance the diversity of 
agent behaviors~\cite{yang2018meanfield} of MARL instead of homogeneous roles, the agents can be grouped selectively~\cite{li2021cds,wang2021rode} 
based on similar trajectories. 
For instance, SePS~\cite{christianos2021seps}
performs a one-shot clustering and reuses weights within each cluster. 
However, a major limitation of these approaches lies
in that the role assignment of agents 
remain largely static --- 
that is, once the  
role (group) is decided, an agent cannot be re-assigned. 
Therefore, these approaches may not adapt to the evolving urban mobility environment.
DyPS~\cite{wang2024dyps} periodically re-clusters agents based on 
latent trajectories (e.g., states, actions, rewards), and support the urban resource allocation.  
However, as each group of agents 
holds its own policy network, 
it remains difficult to further
enhance the scalability of the approach
given large number of agents and groups in practice.

$\bullet$~\textbf{Urban Mobility Resource Allocation.}
Reinforcement learning~\cite{HeShin2022CapsuleRL,YangHeTabatabaie23}, including MARL,
has been explored for bike sharing
resource allocation~\cite{li2022dockless,staffolani2025cabra}, 
traffic light control~\cite{chu2020traffic}, and ride-sharing~\cite{li2020irebalance}. 
For instance, i-Rebalance~\cite{li2020irebalance} studied shared-policy reinforcement learning for city-wide repositioning of 
idle ride-sharing vehicles.   
However, these approaches often rely on heuristic clustering or operate with one shared policy, which limits their adaptability in the complex urban environments.  

$\bullet$~\textbf{Differences from Prior Arts.}
Different from the above studies,
\name{} advances from the following aspects.  
First, we have enabled 
a low-dimensional identify (ID) embedding as well as the model weights
for every agent. This way, we can enhance 
the invidual learnability of the agents~\cite{terry2020ps}.
Second, \name{} provides a dynamic adjustment of agent role (group), 
enabling city-wide mobility resource allocation.

\section{\mbox{Concepts, Problem Formulation, \& Core Designs}}\label{sec:problem}

\subsection{Concepts \& Problem Formulation}

$\bullet$~\textbf{Spatial \& Temporal Discretization.}
Following the prior studies, we discretize the service area into 
a total of $K$ rectangular regions 
that are indexed by  
$\mathcal{K}\!=\!\{1,\dots,K\}$.  
A set of $N$ agents serving as the  mobility resource re-allocators (e.g., a fleet of trucks or coordinators for bike rebalancing) 
is denoted by  
$\mathcal{A}\!=\!\{1,\dots,N\}$ where each agent $i$ serve a region at each
time interval $t$.  
We discretize the time horizon into
time intervals (e.g., days in our current studies).

$\bullet$~\textbf{States \& Actions.}
We design the mobility resource states
for \name{} to capture and learn. 
Specifically, 
for each time interval $t$,
we have a global state $\textbf{s}_t$ which consists of
mobility features (i.e., time of a day, day of a week, 
distributions of available mobility resources, historical pick-up statistics)
and urban environment features (harvested from OpenStreetMap). 
In this prototype study, we encode 
the time of a day ($h_t$ in hour), day of a week $d_t$ (integers from 0 to 6) of an time interval $t$ into 
a vector of 
$[\sin \tfrac{2\pi h_t}{24},\cos \tfrac{2\pi h_t}{24},\sin \tfrac{2\pi d_t}{7},\cos \tfrac{2\pi d_t}{7}]$. 
We take in the availability 
of mobility resources (e.g., number of available bike inventory) per region 
as $[b_t^1, b_t^2, \ldots, b_t^K]$. 
For each region $k$ at the time interval $t$, 
we find the means $\mu_t^k$ and standard deviations $\sigma_t^k$
of aggregate pick-ups (from all stations in a region) in the most recent $H$
time intervals, and therefore
we have for all regions 
the statistics vector $[\mu_t^1, \sigma_t^1, \mu_t^2, \sigma_t^2, \ldots,  \mu_t^K, \sigma_t^K]$. 
For each region $k$, we encode the numbers of roads, bike lanes, and
POIs within the region into a three-dimension
vector $[\text{rd}^k, \text{bd}^k, \text{pd}^k]$~\cite{liu2023gplight}.

For each agent $i$ at time interval $t$, 
we find the sizes 
of mobility resources (e.g., bikes)
that will relocate from
the current region 
to one of the four adjacent regions
(to the north, south, east, and west)
as 
$\mathbf{a}_t^i=
[a^{(\text{North},i)}_t,
a^{(\text{South},i)}_t,
a^{(\text{East},i)}_t,
a^{(\text{West},i)}_t]$. 
The value in $\mathbf{a}_t^i$, 
if negative, 
represents the relocation
from an adjacent region.

Let $d_t^k$ and $o_t^k$ respectively
be the numbers of pick-up requests and drop-offs at the region \(k\in\mathcal{K}\) 
at the time interval $t$.  
Given the states and actions, 
\name{} finds and updates the availability of 
mobility resources for each region
as
\begin{equation}
b_{t+1}^k=\max \left\{b_t^k+ \sum_{j:\,\text{loc}(j)\in \mathcal{N}(k)}a_t^{(\!*, j)} - d_t^k + o_t^k, 0\right\},
\end{equation}
where $\mathcal{N}(k)$
represents the adjacent regions of
a region $k$.

$\bullet$~\textbf{Reward Function.}
Given a coordinated relocation
decision \(\mathbf{A}_t=\{\mathbf{a}_t^1, 
\mathbf{a}_t^2,  \dots,
\mathbf{a}_t^N\}\) from $N$ agents, the availability of mobility resources (inventory) in a region \(k\) 
(before serving the actual pick-up requests)
is
\begin{equation}\label{eq:net-inventory}
\tilde b_t^k
  = b_t^k + 
\sum_{j:\,\text{loc}(j)\in \mathcal{N}(i)}a_t^{(\!*, j)}, 
\end{equation}
where \(a_t^{(\!*, j)}\) is the signed net inflow contributed by agent \(i\). The fulfilled demand is then given by  
\begin{equation}\label{eq:served}
S_t^k
  = \min\!\bigl\{d_t^k,\;\tilde b_t^k\bigr\},
\end{equation}
while the unfulfilled demand
is calculated by 
\begin{equation}\label{eq:unmet}
U_t^k
  = d_t^k - S_t^k
  = \max \bigl\{d_t^k - \tilde b_t^k, 0\bigr\}. 
\end{equation}

We define the reward function for 
each agent $i$ at time interval $k$
as 
\begin{equation}\label{eq:agent-reward}\textstyle
r_t^i \;=\;
\lambda\!\Bigl(1-\frac{U_t^i}{d_t^i+\varepsilon}\Bigr)
        \;-\;
\alpha\,\frac{U_t^i}{d_t^i+\varepsilon}
        \;-\;
\beta\,\frac{\|\mathbf{a}_t^i\|_1}{m},
\end{equation}
where  
the three components ($\lambda, \alpha, \beta >0$) inside the reward function $r_t^i$ respectively 
favor (i) high service ratio, 
(ii) low under-served demand, 
and (iii) low cost of relocation that is proportional to the total size of relocated mobility resources capped by maximum relocation load $m>0$.

$\bullet$~\textbf{Problem Formulation.}
We formulate our mobility resource allocation problem as a finite–horizon multi-agent Markov game.   
For a horizon of $T$ time intervals, and a discount factor $\gamma\!\in(0,1]$, \name{} aims to 
find the MARL parameters $\boldsymbol{\theta}$ that maximize the objective function of
\begin{equation}
J(\boldsymbol{\theta})=
\mathbb{E}
\!\Bigl[
   \tfrac1N\sum_{i=1}^{N}
   \sum_{t=0}^{T-1}\gamma^{t}r_t^i
\Bigr].    
\end{equation}

\begin{figure}[t]
  \centering
  \includegraphics[width=\linewidth]{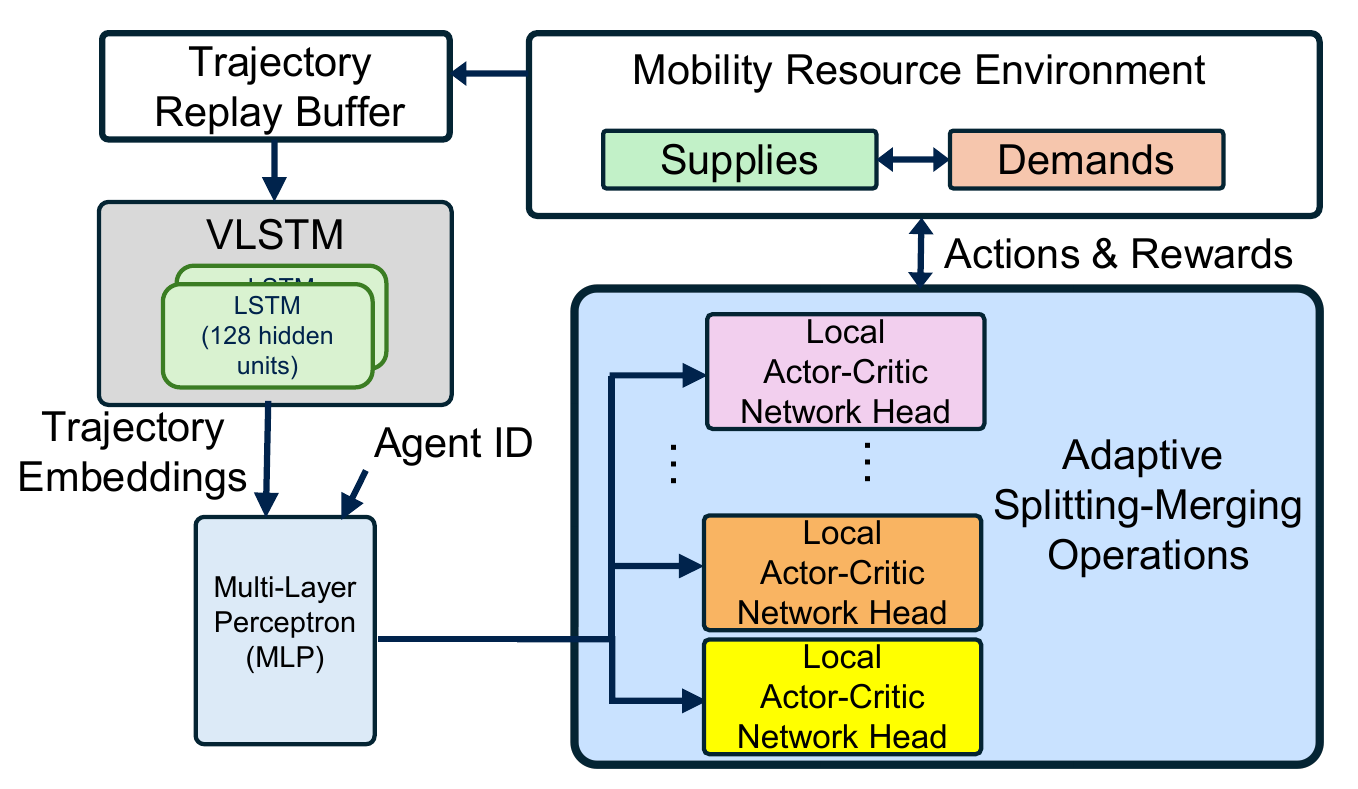}
    \vspace{-4mm}
  \caption{
  Illustration of overall architecture of \name{}. 
    }
  \label{fig:overview}
  \vspace{-4mm}
\end{figure}

\subsection{Hierarchical
Adaptive Grouping}
\label{sec:hagps:arch}

We have designed 
the hierarchical adaptive grouping to dynamically assign
the group (role in mobility resource allocation) for each agent. We have designed adaptive splitting and merging methods, such that the number of groups
and the sizes of each group
dynamically change as the learning evolves. 

To better serve the urban-scale setting, the global group can serve 
as the macro coordination for a district or a county. 
Each global group indexed by $\mathcal{C}_t^{(c)}$ at the interval $t$ has a feature trunk network
$
T_{\boldsymbol{\theta}_c}
$
shared across all the agents inside the group $c$. 
The resulting embeddings 
$
\mathbf{h}_t^i = T_{\boldsymbol{\theta}_c}(\mathbf{s}_t).
$
Within each group, we further partition multiple local groups, each of which
$(c, l)$ maintains 
a local actor–critic network head which has compact structure.
Such local groups can serve
underneath the global group as micro coordination
for the neighborhoods or street levels.
The local actor-critic network
head of the agent $i$ is formed by a multi-layer perceptron (MLP) with 128 units and maps the concatenated
vector of 
$\mathbf{h}_t^i$
and identity embeddings 
$\mathbf{e}_i$
into the action $\mathbf{a}_t^i$ and reward $r_t^i$.

$\bullet$~\textbf{Dynamic \& Adaptive Agent Grouping.}
After every episode, each agent encodes its latest $H$‑step trajectory
\begin{equation}
\boldsymbol{\tau}_{t-H:t}^{\,i}
 =\left\{\mathbf{s}_{t-H}^i,
 a_{t-H}^i,
 r_{t-H}^i,
 \dots,
 \mathbf{s}_{t-1}^i,
 a_{t-1}^i,
 r_{t-1}^i\right\},     
\end{equation}
via a Variational Long Short-term
Memory (\texttt{VLSTM}) with 128 hidden units and obtain
the embeddings, i.e.,
\begin{equation}
\mathbf{z}_t^i = \texttt{VLSTM}\left(\boldsymbol{\tau}_{t-H:t}^i\right). 
\end{equation}

For each local group $l$,  
let $|\mathcal{G}_t^{(l)}|$
be its group size, and
we find the average embeddings
\begin{equation}\label{eq:hagps-centroid}
\mu_t^{(l)} =
  \frac{1}{|\mathcal{G}_t^{(l)}|}
  \sum_{j\in\mathcal{G}_t^{(l)}}\! \mathbf{z}_t^{\,j},
\end{equation}
and the average symmetrized KL divergence within the group, 
\begin{equation}
D_t^{(l)} =
  \frac{1}{|\mathcal{G}_t^{(l)}|}
  \sum_{j\in\mathcal{G}_t^{(l)}}\!
    \tfrac12\bigl(
      \mathrm{KL}(\mathbf{z}_t^{\,j}\,\|\,\mu_t^{(l)})
     +\mathrm{KL}(\mu_t^{(l)}\,\|\,\mathbf{z}_t^{\,j})
    \bigr).
\end{equation}

\textbf{Splitting and Merging Operations.}
The splitting and merging operations of the agents are given as follows. 
If $D_t^{(l)}>D_{\text{split}}$ and
$|\mathcal{G}_t^{(l)}|$ is greater than the pre-defined group size $S_{\min}$, we bisect an agent group $l$  with
$k$‑means ($k{=}2$) over $\{\mathbf{z}_t^i\}$.   
Both local groups inherit the parent's local actor-critic network heads, and immediately
re‑cluster their members into a total of $S_{\max}$ sub‑groups.
Two local groups inside the same global group are merged if
$\mathrm{KL}(\mu_t^{(a)} || \mu_t^{(b)})
+
\mathrm{KL}(\mu_t^{(b)} || \mu_t^{(a)})$
is less than a pre-defined merging threshold
$\tau_{\text{merge}}$.
The local actor-critic network heads of the larger sub-group are kept, while the agents are re‑assigned via
$k$‑means to the nearest of the $S_{\max}$ centroids.

\textbf{Adaptive Regrouping Period.}
In order to enable adaptive regrouping period instead of a fixed one, we find a running exponential average
\(
  \bar D_t
   = \eta\,\bar D_{t-1}
     +(1{-}\eta)\tfrac{1}{|\mathcal{G}_t^{(l)}|}\!\sum_{l=1}^{|\mathcal{G}_t^{(l)}|}\!D_t^{(l)}
\)
to adjust the period before the next split–merge operations, i.e.,
\begin{equation}\label{eq:delta}
  \Delta_{t+1} =
    \max\Bigl(
      1,\;
      \bigl\lceil
        \Delta_0\,e^{-\zeta(\bar D_t-\delta)}
      \bigr\rceil
    \Bigr)\;.
\end{equation}
Thus, when KL divergence within the local group has stabilized, regrouping becomes infrequent.  
As we can bound the 
maximum numbers
of global group and local groups, we can restrict the 
model parameters and subsequent memory footprint
of the feature trunk network 
and the local actor-critic network head. 
Our future extension will 
include detailed theoretical 
analysis over the performance.

\section{Experimental Evaluation}\label{sec:exp}

We present our experimental 
settings, baselines, and experimental results as follows.

\subsection{Experimental Settings}\label{sec:exp:env}

\noindent
$\bullet$~\textbf{Dataset Preparation.}
We leverage a total of 
1,232,838 trips in January 2024 for our experimental studies. 
We have aggregated them into an origin–destination matrix over 
a total of $K = 106$ 1x1 km\textsuperscript{2} rectangular regions covering the central Manhattan.

$\bullet$~\textbf{Comparison Baselines.}
We compare our approach with 
the following baseline approaches. 
\begin{itemize}[leftmargin=*,nosep]
\item \textbf{No-Share}: which coordinates the mobility resources with fully independent proximal policy optimization (PPO).
\item \textbf{Share-All}: which has one global actor–critic shared by all agents.
\item \textbf{SePS} \cite{christianos2021seps}: 
which performs offline $k$-means clustering for agent grouping.
\item \textbf{DyPS} \cite{wang2024dyps}: which performs dynamic grouping with  group networks.
\item \textbf{CDS} \cite{li2021cds}: which performs diversity-regularized full parameter sharing.
\item \textbf{\name{} w/o ID}: which removes the identity (ID) embeddings from \name{}.
\item \textbf{\name{} w/o SM}: which performs no split–merge operation (i.e., fixed group sizes).
\item \textbf{\name{} w/o HG}: which performs no hierarchical grouping.
\item \textbf{\name{} w/o ARP}: which performs no adaptive regrouping period.
\end{itemize}

$\bullet$~\textbf{Implementations \& Detailed Parameter Settings. }
All models are implemented  in PyTorch and trained on a single T4 GPU (16 GB RAM) on Google Colab.
The PPO implementation follows the 37-detail checklist \cite{shengyi2022the37implementation}
(mini-batch SGD, value-loss clipping, etc.).
We configure the model and environment settings to balance performance and training stability. The reward function employs weighting coefficients $\alpha = 5.0$, $\beta = 15.0$, and $\gamma = 3.0$ to respectively emphasize fulfillment of demand, penalize under-service, and discourage excessive relocations. 
Environment inputs include temporal encodings with 4 dimensions, spatial features with 6 dimensions, and one-step demand history represented by a single dimension. 
Each training epoch consists of 64 episodes (corresponding to 64 simulated months), and each episode spans a maximum of 31 decision steps to cover the days in January. 
We apply PPO with a discount factor $\gamma = 0.995$ and generalized advantage estimation (GAE) with $\lambda = 0.95$. The learning rates for policy and value networks are set to $\texttt{3e{-}4}$ and $\texttt{1e{-}3}$, respectively to reflect a more conservative update for the policy to ensure stable learning while allowing faster adaptation of the value function.

To ensure robust evaluation, we reserve 20\% of the training data for validation purposes.
The split–merge controller is governed by five stable hyper-parameters.  
The regroup interval is initialized to $\Delta_0 = 8$ episodes and clipped to the range $\Delta \in [1,\,64]$.  
After each episode the running KL divergence is updated with an exponential smoother ($\eta = 0.90$) and compared with the target drift level ($\delta = 0.02$); any excess shortens the next interval with sensitivity $\zeta = 3$.  
This configuration balances rapid reaction to behavioral shifts against the computational overhead of regrouping.

\smallskip
\noindent\textbf{Evaluation  Metrics.}
Following the prior practices in bike resource reallocation \cite{foerster2016ldr,staffolani2025cabra},
we evaluate overall performance via (1) fulfilled service ratio, i.e.,
\begin{equation}
\mathrm{Avail}_T \;=\;
1\;-\;
\frac{\displaystyle\sum_{t=0}^{T-1}\sum_{k=1}^{K}U_t^k}
     {\displaystyle\sum_{t=0}^{T-1}\sum_{k=1}^{K}d_t^k},
\end{equation}
i.e.\ the fraction of all pickup requests that are successfully served over an episode; and (2) the total number of bikes that get rebalanced. 
Higher fulfilled service ratio 
and total bike rebalanced indicates better performance 
in mobility resource allocation.

\subsection{Preliminary Experimental Results}\label{sec:exp:performance}

Table \ref{tab:main} summarizes the fulfilled service ratio and the total number of bikes that get rebalanced. We can see that {No‑Share}, which is the fully‑independent PPO baseline, achieves only about 51 \% fulfilled service ratio, demonstrating the drawback of over‑parameterization without sharing the model parameters. Share-All reserves the model parameters but its performance falls below 44\%. Compared with CDS, SePS, and DyPS, our achieves a higher fulfilled service ratio of 77.21\% and more bikes rebalanced thanks to the gained learnability from its hierarchical adaptive grouping-based parameter sharing.

We also demonstrate the ablation studies within Table \ref{tab:main}. We can see that removing the identity (ID) embeddings degrades about 0.3 percent in terms of fulfilled service ratio. This indicates that a small agent‑specific vector can help disentangle similar mobility allocation policies. 
Disabling the splitting and merging operation or 
the hierarchical adaptive grouping leads to more performance degradation, 
and decreases fulfilled service ratios by about 2.1 percent and 4.0 percent, respectively. 
This underscores their importance for the mobility resource allocation.  
Fixing the regrouping period instead of an adaptive one
leads to 1.1 percent performance drop in terms of fulfilled service ratio. 
Combining all these designs lead to overall superior performance of \name{}.

\begin{table}[t]
  \caption{
  Performance comparison
  of different schemes. 
  }
  \label{tab:main}
  \footnotesize
  \resizebox{\columnwidth}{!}{%
  \begin{tabular}{lcc}
    \toprule
    Method & Fulfilled Service Ratio (\%) & Total Bikes Rebalanced\\
    \midrule
    No‑Share                  & 51.18 & 357,864 \\
    Share‑All                 & 43.84 & 333,372  \\
    CDS                       & 58.40 & 407,316  \\
    SePS                      & 64.77 & 453,180  \\
    DyPS                      & 69.09 & 462,696 \\
    \rowcolor{gray!12}
    \name{}\ w/o ID           & 76.91 & 471,900    \\
    \rowcolor{gray!12}
    \name{}\ w/o SM  & 75.10 & 470,028    \\
    \rowcolor{gray!12}
    \name{}\ w/o HG    & 73.23 & 468,936    \\
    \rowcolor{gray!12}
    \name{}\ w/o ARP   & 76.07 & 471,588    \\
    \rowcolor{green!15}
    \textbf{\name{}}          & \textbf{77.21} & \textbf{472,212}\\
    \bottomrule
  \end{tabular}}
\end{table}

\section{Conclusion \& Future Work}\label{sec:conclude}
We study in this work a multi-agent reinforcement 
learning named hierarchical and
adaptive 
grouping-based 
parameter 
sharing (\name{})
for dynamic mobility resource allocation. 
Using the NYC bike sharing
as a case study, 
\name{} addresses two challenges regarding MARL for mobility resource allocation -- that is, 
adaptive sharing of policy across agents, 
and memory-efficient parameter sharing 
in the urban-scale setting. 
we have designed a hierarchical approach that consists of
global and local information of the mobility resource states. We have developed an adaptive
budget-capped agent grouping approach to split or merge
the groups of agents based on their relative closeness of encoded
trajectories.
Extensive experimental studies 
based on over 1.2 million bike sharing trips have validated
the performance of \name{}
in rebalancing the demand and supply in a metropolitan setting.
Our future work will include:
(i) expansion of experimental studies; 
(ii) introduction of multi-city data evaluations.

\begin{acks}
This project is supported, in part, by the National Science
Foundation (NSF) under Grants 2239897 and 2303575, and the Connecticut Division of Emergency Management \& Homeland
Security (DEMHS) Hazard Mitigation Grant Program (HMGP). Any opinions, findings, and conclusions or recommendations expressed in this material are those of the authors and
do not necessarily reflect the views of the funding agencies.
\end{acks}

\bibliographystyle{ACM-Reference-Format}
\bibliography{sample-base}

\end{document}